\documentclass{article}

\usepackage[position,final]{neurips_2026}

\usepackage{microtype}
\usepackage{subcaption}
\usepackage{tabularx}
\usepackage{booktabs}
\usepackage{graphicx} 

\usepackage{multirow}
\usepackage{enumitem}

\usepackage{amsmath}
\usepackage{amssymb}
\usepackage{mathtools}
\usepackage{amsthm}
\usepackage{placeins}
\usepackage{algorithm}
\usepackage{algorithmic}

\usepackage{wrapfig}

\newcommand{\eat}[1]{}

\usepackage[capitalize,noabbrev]{cleveref}

\theoremstyle{plain}

\theoremstyle{definition}

\theoremstyle{remark}

\newcommand{\yrcite}[1]{\citeyearpar{#1}}

\title{Adversarial Risk in the Generative AI Era Necessitates Dropping the Small Epsilon Ball}

\author{%
  Andrew C. Cullen \\
  University of Melbourne \\
  \texttt{andrew.cullen@unimelb.edu.au} \\
  \And
  Neil G. Marchant \\
  University of Melbourne \\
  \And
   Paul Montague \\
   DST Group, Adelaide \\
  \AND 
    Jiani Xie \\
  University of Melbourne 
   \And
  Benjamin I. P. Rubinstein \\
  University of Melbourne \\
}

\begin{document}

\maketitle

\begin{abstract}
Adversarial Machine Learning research is currently centered on the search for the epsilon-ball: minimal perturbations designed to fool models while remaining imperceptible to humans. We argue that this fixation on $\ell_p$ norms is a metrological trap that communicates a distorted perception of risk to defenders, one that is unaligned with real-world system vulnerabilities. This paradigm misleads defenders into optimizing for narrow, mathematical robustness at the expense of systemic security, while simultaneously incentivizing attackers to exploit non-overlapping alternative adversarial pathways. Ultimately, we argue that the dominance of the $\ell_p$ ball does not just fail to reduce harm---it may actively facilitate it by communicating a false sense of security against the wrong threats.

\end{abstract}

\section{Introduction}

The field of Adversarial Machine Learning (AML) has centered itself around a dominant paradigm: the epsilon-ball. Ever since it was discovered that neural networks can be subverted by small perturbations \citep{szegedy2013intriguing}, the community has continually explored risk through the lens of small, distance-bounded perturbations (most notably the $\ell_p$-ball). 
This framework is advantageous, in that it aligns well with the mechanics of machine learning \citep{goodfellow2014explaining, madry2017towards}, while producing abstractions that are proxies for real trade-offs for learning in adversarial environments. 

However these abstractions may not be well aligned with risk in real environments. After all, if these attacks were as simple and powerful as research portrays, then the AI era should have led to a wide scale proliferation of such manipulations. Within this work we explore the counterintuitive argument that the field's attention on these abstractions may have led to a false sense of security, blinding us to broader research opportunities, and that fundamentally it may be driving an understanding of risk that may well be leaving systemic vulnerabilities under-addressed.

Specifically, we argue that the epsilon-ball paradigm fails us through four critical axes:
\begin{enumerate}[leftmargin=*, noitemsep, topsep=0pt]
    \item \textbf{The Metric‑Alignment axis:} Metric proximity in input space is considered to be a key measure of risk. However, this assumption fails across modalities: tiny pixel‑space changes may alter semantics, while large textual or audio edits may preserve them~\citep{jia2017adversarial, wei2023jailbroken}. The problem is not merely that epsilon-balls are imperfect perceptual proxies; it is that no fixed input‑space metric adequately captures the vulnerabilities of modern ML systems.    
    
    \item \textbf{The Ecological axis:} The precepts of lab-optimized attacks often collapse when considered within real-world conditions~\citep{zheng2024physical}, especially for systems outside the traditional computer vision domain. This leads to a phenomenon that we dub the \emph{Empty Room Paradox}. 
    \item \textbf{The Technological axis:} Generative AI enables attackers to create arbitrary, high‑fidelity synthetic inputs that lie outside any notion of a small perturbation~\citep{dai2024advdiff}. The community has explored many such attacks empirically, but these capabilities remain largely unformalized, leaving a gap between practical risk and current theoretical abstractions. 
    \item \textbf{The Incentive axis:} Ignoring incentives for attackers abstracts away the very reasons for performing an attack~\citep{ehrenberg2025adversaries}. Similarly, by presenting a view of risk driven by the epsilon-ball, we incentivize attackers to work in other spaces, further minimizing utility.
\end{enumerate}

By consolidating these arguments, we propose that the productive frontier of AML should move from mathematical optimizations to an articulation of systemic safety. \textbf{In this position paper, we argue that the tools through which we study risk in AML are inherently biasing our understanding of risk.} We argue for a fundamental pivot from abstract games of adversarial hide and seek between attackers and defenders in (oftentimes) pixel space toward a broader view of risk. In this framework, the realistic perception of risk is defined by the socioeconomic intent of the adversary, the provenance and transformation pipeline of input data (like sensor data), and the production constraints of the deployment loop. This position paper serves not as a simple critique of research practices, but as a \textbf{call for action to acknowledge that the metrological bounds of current research practices are shaping an inaccurate portrayal of risk.} Correcting this is crucial for understanding the conditions that drive safety, risk, and harm in the real-world.

\section{The Orthodoxy of the Epsilon-Ball}\label{sec:setup}

Small metric perturbations are an appealing mathematical abstraction of risk, which links risk and detectability with the size of the perturbations~\citep{gilmer2018motivating}. However, in aligning with this paradigm, subsequent research has become predicated upon a shared set of assumptions: that risk is atomized to particular pathways, that stealth is necessary, that deployment details can be abstracted away, 
and that minimizing the distance to benign sources is an optimal attacker strategy. While it is true that the community does consider a broad suite of threat models, Table~\ref{tab:modality_clean} demonstrates that the $\ell_p$ epsilon-ball paradigm is the most dominant, especially when further interrogation of the Poisoning, Semantic/Transformation, and Physical categorizations are also dominated by the epsilon ball, with metrics relying upon bounding small perturbations (see Appendix~\ref{app:experimental}). While the soft constraints of Jailbreaks are an exception, they still reflect the orthodoxy of the small perturbation paradigm.

\begin{wraptable}{r}{0.5\textwidth}
\vspace{-25pt} 
\caption{Threat frequency from $4311$ Adversarial ML papers from ICML/ICLR/NeurIPS over 2022--2025, categorized by primary threat model (all entries are distinct) following Appendix~\ref{app:experimental}. Abbreviations: $\epsilon$ : small metric perturbations, Pois.: Poisoning, Jail.: Jailbreak, Sem.: Semantic, Def.: Defense. Note: \textbf{the majority of Poisoning, Jailbreaking, Patch, and Defense papers employ an $\mathbf{\epsilon}$ threat model} (see  Section~\ref{sec:setup} and Appendix~\ref{app:experimental}). } 
\label{tab:modality_clean}
\centering 
\begin{small}
\begin{sc}

\setlength{\tabcolsep}{4pt} 
\begin{tabular}{lrrrrrr}
\toprule
Category & $\epsilon$ & Pois. & Jail. & Sem. & Patch & Def \\
\midrule
Audio      & 11   & 18   & 25   & 9    & 2   & 1   \\
General    & 293  & 118  & 98   & 114  & 13  & 28  \\
Graph      & 27   & 30   & 20   & 14   & 5   & 9   \\
Image      & 334  & 153  & 143  & 110  & 31  & 51  \\
Multim. & 60   & 38   & 43   & 23   & 4   & 5   \\
RL         & 46   & 27   & 51   & 43   & 11  & 20  \\
Text/LLM   & 134  & 303  & 349  & 75   & 14  & 30  \\
Unknown    & 75   & 16   & 48   & 34   & 4   & 11  \\
Video      & 20   & 15   & 27   & 15   & 7   & 3   \\
\midrule
Prop. ($\%$)     & 31 & 22  & 25  & 14  & 3  & 5 \\
\bottomrule
\end{tabular}
\end{sc}
\end{small}

\vspace{-15pt} 
\end{wraptable}

\subsection{Images: Pixel-Level Perturbations}

AML has historically primarily considered Computer Vision problems, where risk is typically modeled as a failure to correctly label an image under small $\ell_p$ variations. These attacks typically employ the very gradient-based mechanisms that are so crucial for AI learning~\citep{goodfellow2014explaining, madry2017towards, carlini2017towards}. This has produced an arms race between attackers and defenders, where stealth is correlated with perturbation size~\citep{goodfellow2014explaining, cohen2019certified, cullen25a}.

But within this arms race, there are a common set of rules that are often followed, which do not necessarily align with real world risks. Digital-only attacks assume that the attacker has digital access to an input~\citep{goodfellow2014explaining}, which suggests a level of systemic compromise that may obviate the need to adversarially attack. For physical world attacks involving introduced artifacts~\citep{kurakin2016adversarial, brown2017adversarial, eykholt2018robust} it is often assumed that environmental conditions are constant, however performance is inherently subject to lighting, camera noise, and viewpoint changes that can dilute the impact of the adversarial effect~\citep{lu2017no, athalye2018synthesizing, jan2019connecting}. These small perturbations do map to a real-world phenomenon, in that sensors and compression techniques used by digital cameras and scanners can introduce digital artifacts that can induce deleterious model performance.

However, the extension of this logic to adversarial perturbations is not necessarily clear. To explain this, consider \textbf{who is being tricked by an adversarial perturbation}. On one level, the answer to this is simple, it is the model. But if we are tricking the model in isolation, then the size of the perturbation is immaterial, as any change is enough. Thus if the paradigm is to be meaningful, the model must exist within a broader framework, in which the epsilon-ball also matters to either a human or machine detector. While a human detector exposed to the input and the models output may detect the adversarial perturbation, as we will discuss in subsequent sections such processes (a) fail at the scales associated with AI workflows and (b) are at risk of failure in the face of automation bias. This is especially so when attacks can always learn to avoid or overwhelm the pipeline for human review, or to rely on interventionist complacency~\citep{parasuraman2010complacency, horowitz2023bending}.

If we rely upon machine detection, it is not clear that aligning the risk model to an epsilon $\ell_p$ ball leads to improved machine detectability. In fact, tying machine detectability to simple epsilon-balls fails to acknowledge the history of statistical methods and transformation based defenses \citep{nesti2021detecting, sitawarin2022demystifying} detecting off-manifold samples with high probability, and detecting adversarial examples may be just as hard as designing models robust against them~\citep{tramer22a}. 

The emergence of generative architectures further destabilizes this paradigm. Traditional AML considers risk and detectability in terms of the $\ell_p$ norm of an additive noise vector $\delta$---leading to perturbations that take the form of high-frequency artifacts. However generative models can synthesize photorealistic, semantically adversarial samples from scratch~\citep{dai2024advdiff}, bypassing the need to introduce detectable high-frequency pixel artifacts. In this regime, attacks evade detection not because the noise is mathematically small, but because the synthesized image is statistically natural, making the arms race over pixel-budgeted stealth effectively obsolete.

\subsection{Audio: Psychoacoustics and Physical Constraints}\label{sec:audio}

Audio data leads to a broader diversity of modeling paradigms---ranging from Automatic Speech Recognition (ASR) to Speaker Verification, diarization\footnote{Diarization is the task of segmenting a recording into speaker-spoken intervals.}, and instruction-following models, that have led audio to become a leading modality of AI-enabled human computer interaction. Our understanding of the risk facing these systems is built upon foundational work by Carlini \& Wagner \yrcite{carlini2018audio}, setting the stage for perturbations that can transcribe, and potentially action, malicious commands~\citep{du2020sirenattack, li2020advpulse, yu2023smack}. 

Audio, by its very nature, provides a unique illustration of the limitations of perceptual bounds. While many works consider the average Signal to Noise (SNR) ratio, which is conceptually analogous to a scaled small-epsilon $\ell_p$ ball, extensions have been made to consider more perception tied measures, through by exploring the concept of psychoacoustic risk. This framework leverages models of human acoustic acuity to introduce stealthy adversarial perturbations that align with the parts of the audible domain that human perception is likely to perceive~\citep{abdullah2019hear, abdullah2021sok}. Adversarial perturbations have also been hidden within music or other forms of noise \citep{schonherr2018adversarial, abdullah2021beyond}. More extreme threat models include inaudible commands like DolphinAttack \citep{zhang2017dolphinattack}, in which voice commands are modulated onto ultrasonic frequencies detectable only by hardware sensors, bypassing audibility entirely.

However, when it comes to the implementation of these attacks, the devil is still very much in the details. Real-world attacks induce significant throughput concerns, incentivizing digital-domain attacks. But these attacks inherently abstract away the physics of acoustics on the precise perturbations constructed, with reverberation, background noise, nonlinearities in recording and transmission devices can also the attack beyond its successful envelope~\citep{yakura2018robust, schonherr2019imperio}, with real-world attacks needing to cross an energetic barrier for success. However this necessitates louder perturbations to survive transmission, which decreases the difficulty of detection \citep{chen2020metamorph}.  

While real-time and real-world attacks do exist \citep{oreilly2022voiceblock, chiquier2022real, raina2024muting, xie2026hearing}, attack success rates (ASR) drop precipitously, with past works observing ASR's above $90$ in digital offline settings, with the same attacks struggling to breach $45$ in over-the-air attacks, when room reverberation and background noise are factored in~\citep{chiquier2022real}). In practice, reliably attacking models using current methodologies requires attack amplitudes at a level for which stealth is impossible.

The unreality of metric-bounded acoustic risk is amplified by the rise of zero-shot voice synthesis and generative deepfakes. Rather than hiding adversarial perturbations beneath psychoacoustic masking thresholds, an adversary can clone an identity from a brief reference prompt~\citep{wang2023neural, le2023voicebox}. This shifts the threat from perturbations to \emph{de novo} synthesis, in which generative approaches bypass measures of detectability. An attacker that has the capacity to successfully attack a system without the expense and difficulty of a perturbative attack will likely never pursue such a pathway, rendering many acoustic AML approaches obsolete.

\subsection{Discrete Semantic Shifts: Text, Code, and LLM Risk}

The ubiquity of natural language interfaces for AI systems, including discrete modalities like code generation and malware detection, makes it crucial for us to investigating the robustness of such systems. Initial adversarial text works adapted the continuous $\ell_p$ framework into discrete analogues: character-level via homoglyphs and pseudo-typing mistakes; word-level through synonym substitution via methods like TextFooler \citep{jin2020bert}; token-level suffix optimization~\citep{zou2023universal} and sentence-level through semantic paraphrasing. These works frame risk as an optimization problem constrained by soft distance metrics like edit distance or perplexity bounds, under the assumption that an attack must preserve the text's underlying semantics to an observer while inducing misclassification. 

However, modern language models fundamentally break this perturbation model. In both direct and indirect prompt injection~\citep{greshake2023not}, untrusted inputs hijack an agent's execution. The attack payload is not a perturbed variant of a benign document, but rather it is a semantically coherent set of instructions that exploits the model's inability to separate control instructions from data. Similarly, alignment jailbreaks bypass guardrails through hypothetical scenarios, persona adoption, fictionalization, or multi-turn persuasion~\citep{wei2023jailbroken, chao2025jailbreaking} that exhibit normal discourse, and imperceptible metric differences. 

Fundamentally, detection frameworks that rely on perplexity thresholds or token density estimation are calibrated to catch unnatural sequences, ignoring threats encoded in fluent natural language. Moreover, a focus on human perceptibility ignores the fact that most text systems operate at a scale far beyond human-in-the-loop analysis. This has motivated the machine detectability community to consider metrics beyond $\ell_p$-style proxies~\citep{ding2024vulnerability, zhou2025large}. The combination of this misalignment between communities, and the power of synthetic text generation (see Section~\ref{sec:generative_ai} for more detail) further dilutes the validity of small perturbations as the arbiter of risk.

This disconnect is further magnified in Vision-Language Models (VLMs), where threats can bridge both continuous and discrete domains~\citep{yang2026semantic}. Effective multimodal exploits frequently rely upon visual typographic attacks---such as handwritten instructions in a camera feed---which can introduce large $\ell_p$ pixel displacements while functioning as semantically pristine prompt injections.

\subsection{Graphs: Topological Budgeting}
Graph Neural Network (GNN) risk is modeled as the manipulation of graph structure (adding/deleting edges) or node features. The \textit{threat model} assumes an attacker can create fake nodes (eg. accounts) or compromise existing nodes to influence a recommendation or fraud detection system. The \textit{perturbations} are often constrained by a topological budget that allows only $k$ edge modifications \citep{zugner2018adversarial}, with attacker cost (sometimes considered to be a proxy for stealth) considered proportional to the number of changes, or preservation of global graph statistics.

Superficially, this serves as the strongest case for the utility of the small perturbation threat model. However, one must consider how detectable added nodes are to the kinds of graph networks of interest. Social and influence networks can scale to billions of nodes, with huge numbers of nodes being naturally added every day. Under such circumstances, even large $k$ edge modifications can blend into the crowd, suggesting that risk is contextual, rather than absolute. 

\subsection{3D Data: Spatial Shifting and Phantom Objects}

For 3D point clouds and LiDAR-based perception, risk involves shifting point coordinates or additional data points \citep{xiang2019generating}. This threat model is heavily aligned with autonomous vehicles using LiDAR for object detection, as an adversarial scene could trigger real-world impacts, with sensor failure triggering---or failing to trigger---emergency breaking \citep{cao2019adversarial}. This modality again considers an analogue of the $\ell_p$ ball, by way of the Chamfer or Hausdorff distances. However, to cause a geometric effect in LiDAR, the attacker must make large, structured, and often obvious changes to the environment~\citep{kobayashi2025invisible, tu2020physically}, invalidating the concept of small perturbations. Moreover, the inherently sparse and noisy nature of LiDAR data often results in small distance perturbations manifesting as physically impossible object placements---such as points floating in free space without support \citep{hu2023pointca, xiang2019generating, kim2021minimal}.

Discussions regarding metric relativity and reinforcement learning can be found in Appendices~\ref{app:relativity} and \ref{app:rl}.

\subsection{The Common Thread: Perceptual Paradoxes and Metric Traps}

All of these diverse modalities share a common orthodoxy, in that it is assumed that metric proximity is a proxy for security. However, this framing rests on foundational metrological contradictions, that compromise the applicability of AML research to real world problems. 

To explore this, we introduce what we dub the \textbf{Empty Room Paradox}. In physical domains like audio or vision, threat models involve minimizing a metric to ensure perturbations remain imperceptible to human observers. Yet the physical deployment of these attacks renders such stealth illusory. In audio, attackers listen to and record $x$, before adding a perturbation $\delta$ and then broadcasting $x + \delta$ \citep{yakura2018robust, schonherr2019imperio}. In physical vision,  attacks against autonomous vehicles or surveillance rely upon shining projectors onto road signs~\citep{lovisotto2021slap}, or aiming lasers at cameras~\citep{duan2021adversarial}.

However, in both modalities, the deployment mechanism itself will always be glaringly obvious. In audio, the rebroadcast of $x + \delta$ will always be heard as a secondary rebroadcast. Irrespective of how psychoacoustically imperceptible the perturbation is, this secondary rebroadcast will be instantly detectable. Similarly, a vision attack that involves a laser or a projector will serve as a glaringly obvious attack vector, irrespective of the impact on the final object. Thus these threat models implicitly assume that the attack occurs in an \emph{empty room} devoid of any observers. This renders the constraint of human imperceptibility meaningless. And if the room is empty, then the attacker could simply broadcast an overt acoustic command or physically obscure the camera lens. 

This same paradox also extends to textual data. If inputs are automatically processed without human review, constraining an attack to subtle token substitutions is an artificial limitation. And if a human reviewer is present, then. As such, the Empty Room Paradox should be employed as a broader framework for evaluating the realism of risk against deployed AI systems. 

This observer paradox is one of several metrological traps that plague the AML research space, including:
\begin{enumerate}[leftmargin=*, noitemsep, topsep=0pt] 
\item \textbf{Unknown and Non-Stationary Manifolds:} Defining proximity requires an anchor to a safe data manifold, which cannot be known in production systems. Even if it is approximated, it ignores the potential for distributional drift, which would render static geometric boundaries unworkable.                                                              \item \textbf{Abstracted Access Detectability:} Focusing on minimizing the size of perturbation ignores the detectability of the query trajectory required to find it. Pattern inspection can detect iterative optimization patterns long before an attack converges~\citep{li2022blacklight, chen2020stateful}.                                         
\item \textbf{Incentive Distortion as Security Theater:} By treating local metric balls as the primary risk, defenders are incentivized to invest in defenses against what is often the most expensive and detectable attack pathway. As we will discuss in the subsequent sections, practical adversaries will follow the path of least resistance, eschewing $\ell_p$ in favour of generative synthesis, social engineering, or supply-chain compromises.                    
\end{enumerate}    

While $\ell_p$ norms---and the certified guarantees built upon them~\citep{cohen2019certified, cullen2022double, cullen2024s, cullen2026again, cullen2026halt}---provide a mathematically convenient model for analyzing the geometry of local representations, they cannot serve as a comprehensive proxy for end-to-end security in operational systems. Measuring the susceptibility or resistance to variations in local spaces provides no defense against adversaries that can influence systems through unconstrained semantic or generative pathways.

\section{Deployment Loops: Perceptions vs Reality}

Emphasizing problems with technical solutions inherently abstracts away model deployment. However, as has been demonstrated in traditional cybersecurity, risk is concentrated in the how and where or system deployment. This is not to say this space has not been explored, with Carlini et al. demonstrating that models can be hijacked using expired URLs referenced in major datasets~\citep{carlini2024poisoning} and that LLMs can be tricked into releasing their training data or weights~\citep{carlini2021extracting}. These intersect with both established research into the risks of 
code sharing and generalised software supply chains~\citep{ohm2020backstabber, duan2021towards, ladisa2023}, which is crucial given the risk of platforms sharing pre-trained models~\citep{jones2024we, hfhubdocs2025, pytorchhubdocs2025}, which has lead to attacks against AI models~\citep{jiang2022empirical, casey2024large, wang2025sok, meikle2025}. These supply chain risks are exacerbated by the high degree of centralization in AI systems. Beyond foundation models and datasets, the very infrastructure of learning---such as PyTorch and Tensorflow--- including remote code execution~\citep{CVE2025a, CVE2025b, CVE2025f}, denial of service~\citep{CVE2025c, CVE2025f}, and overflow errors~\citep{CVE2025d}. In general, the majority of identified vulnerabilities relate to input validation, overflow, and memory corruption~\citep{cvedetail_pytorch, cvedetail_tensorflow}. However, these works are the exception, rather than the rule, and have done little to shape the community consensus on risk. The dependence of most AI work upon these core tools and common workflows presents an infrastructure bottleneck that we argue poses a far more significant source of risk than any epsilon-ball bounded AML attack.

\subsection{The Economic Gap: Adversary Incentives}

The concentration of deployed systems strongly influences the economics of adversarial manipulation. If deployed models were unique, the investments required to conduct a successful attack might preempt all bar the most motivated attackers. However, this calculus changes when this infrastructure can be re-used across attacks against multiple systems sharing core commonalities. This is especially true in the foundation model era, where multiple systems can be built upon a single foundation model, a successful attack can potentially succeed against multiple systems, reshaping the risk calculus. This vulnerability also relates to shared datasets, as a poisoning attack deployed against a commonly employed dataset will propagate the vulnerability to thousands of downstream applications~\citep{jia2022badencoder, carlini2024poisoning}. While AML researchers may understand this conceptually, our accepted research practices are poorly aligned with such threats, since inference‑time epsilon‑ball defenses offer no protection to these risks.

Fundamentally, the relationship between attackers and defenders is not abstract, but rather exists as a part of a broader game of incentives. If we assume that attackers are rational economic actors optimzing for return on investment, then attackers should be considered not as independently choosing to attack the model, but rather choosing to attack the model because it yields the highest return. But under this model, an AML attack is something that requires an investment in time, data access, and infrastructure. But if a phishing email costs several dollars, works 10\% of the time, and produces the same outcomes as an AML attack, how likely would a rational adversary follow the pathway outlined in the AML literature?

Similarly, an attacker who has access to the model weights likely has the ability to perform far more damaging manipulations than crafting an epsilon small perturbation, as they must have compromised the underlying model. Likewise, an attacker that can support tens of thousands of access requests without being detected could conceivably perform a Distributed Denial of Service (DDoS) attack, which are both simpler to deploy, and require less physical hardware to support than an adversarial attack. 
While it is true that a white-box access model can be a viable platform for developing insight into the worst case dynamics that could affect a model, it is still important to contextualize how we transfer insights from these attacks into a real appreciation of the risk facing deployed systems.

Under this economic lens, it is clear that the most important metric is likely not the size of an $\ell_p$ ball, but rather \textbf{Cost-per-success}, or \textbf{Impact-per-invested-dollar}. Considering $\ell_p$ as the primary threat vector may be elevating the most  \textit{expensive} and \textit{detectable} systemic vulnerability (especially when factoring in query counts), when generative attacks of social engineering may achieve the same outcomes at a fraction of the cost.

\subsection{Machine-in-the-Loop vs. Human-in-the-Loop}\label{sec:HITL}

As Section~\ref{sec:setup} introduced, we must interrogate the undocumented assumptions regarding who or what is being deceived when an adversarial attack is deployed. Epsilon-ball imperceptibility measures typically imply a human-in-the-loop (HITL), as the measure is considered a proxy for human perception. This implicitly frames the ultimate target of the attack as \emph{both} the human verifier \emph{and} the underlying model. However, this assumption collapses under the weight of production scale. In modern MLOps pipelines, data volume vastly outstrips human verification capacity~\citep{eken2025multivocal}, and most deployed systems---like autonomous vehicles, content filters, anomaly detectors, or fraud detectors---produce content never seen by human eyes. Thus there exists automation loops where human imperceptibility is immaterial, and where these machine-in-the-loop defenses alter adversarial incentives regardless of whether a human observer is present. 

The only relevant metric is whether the system's \textit{automated} outlier detectors can perceive the anomaly---which can be a challenge when detection systems are more successful in capturing outliers rather than adversarial intent~\citep{pendlebury2019tesseract}. Of course, such validation requires that systems designers understand this---which may not be clear from the way that the AML community communicates risk.

Even in human-machine teaming efforts, safety cannot be considered a binary function of pixel visibility. Human operations have repeatedly been shown to become complacent and succumb to Automation Bias when paired with intelligent agents~\citep{parasuraman2010complacency, horowitz2023bending}. 

Under such circumstances, an attack does not need to be mathematically invisible to succeed; it only needs to be plausible enough to slip past an overloaded or complacent operator.

Ultimately perceptibility---whether human or machine---is only a valid security metric if 
(a) there is a system in place to perform the perception, and (b) that system is calibrated to an ecological measure of risk. Current defenses fail on both counts. We lack the manpower to watch the data in systems of interest, and 
automated tools are calibrated to detect statistical abnormalities rather than semantic malice. Until our models of risk align with real systems, we will continue to optimize for threat models that exists only in the laboratory.

\section{The Generative Shift}\label{sec:generative_ai}

The growth of generative foundation models fundamentally invalidates the geometric premises that underpin much of adversarial machine learning. Across computer vision, speech, and language we observed that modern generative threats consistently break the small perturbation mold. At its core, classical AML has assumed that adversarial risk is concentrated on local additive perturbation vectors $\delta$, where the size of $\delta$ correlates with risk. These perturbations invariably push inputs off the manifold into low-density decision voids, where $x + \delta$ does not exist within the natural data manifold $\mathcal{M}$. That $x + \delta \notin \mathcal{M}$ underpins much of the literature on detection mechanisms~\citep{stutz2019disentangling, sitawarin2022demystifying}. 

However, generative models can create threats through photorealistic images synthesis, zero-shot voice cloning, fluent LLM prompt injections, deepfakes~\citep{masood2023deepfakes} and more, without ever accessing a benign anchor $x$. Instead the adversary samples directly from an expressive model that approximates the true data distribution to produce outputs $g$ such that $g \in \mathcal{M}$, invalidating the fundamental premise of distance bounded defenses and $\epsilon$-balls. 

By restricting our mathematical formalisms to off-manifold $\ell_p$ perturbations, AML has perfected solutions for a threat model that generative adversaries can bypass with ease. Deployed systems are no longer merely vulnerable to metric corruptions of trusted inputs; instead they face the malicious fabrication of reality itself through samples that challenge our epistemic certainty in  observed data~\citep{ghiuruau2024distinguishing}.

\section{A Framework for Systemic Risk}

Risk is not a geometric property of data, rather it is a \textit{causal property of systems}. A system is at risk not when its decision boundary is crossed by a perturbation of size $\epsilon$, but when an adversary can reliably trigger a harmful downstream consequence. While this approach is certainly considered by the broader information security community, it is clear that a seismic shift is required in AML research practices to align it with real-world risk and harm.

This is a messy truth: perfect mathematical robustness is neither necessary nor sufficient for systemic safety. A model can be mathematically robust ($\ell_p$-stable) yet systemically fragile (vulnerable to supply chain poisoning or semantic manipulation). Conversely, a model can be mathematically brittle yet systemically safe (protected by robust provenance checks, rate limiting, and human oversight). 

Ultimately, the nature of risk is not static, nor is it contained within the weights of a neural network. It is a dynamic property that emerges from the interplay of incentives, access, and deployment constraints. We must step beyond current practices and embrace a more holistic -- if less metrologically convenient -- definition of safety and risk.

\subsection{Harm as a Subset of Risk}

To respond to our proposed need to change the ground rules of AML, we argue that security should be instead considered through the following \textbf{Systemic Risk Framework}:

\begin{enumerate}[leftmargin=*, noitemsep, topsep=0pt]
    \item \textbf{Ecological Realizability:} Can the attack succeed in a noisy, physical, and non-static environment? (Moving beyond digital-only assumptions).
    \item \textbf{Cross-Perceptual Visibility:} Is the attack detectable by \textit{either} a human \textit{or} a machine auditor? (Moving beyond $\ell_p$ imperceptibility).
    \item \textbf{Distributional Authenticity:} Can the system distinguish between a sensor-captured event and a generative model output? (Moving beyond the epsilon-ball).
    \item \textbf{The Harm Function:} What is the socioeconomic impact if the attack succeeds? (Moving beyond the model error).
\end{enumerate}

Thus \textbf{Harm} must be viewed as the downstream consequence of a realized risk. While the AML community has traditionally focused on the \textit{existence} of adversarial examples (a technical vulnerability), we argue that a realistic articulation of risk requires mapping these findings to their resulting societal impact---be it financial loss, physical injury, or the erosion of institutional trust \citep{shapira2025frame}, with perceived impacts magnified in cases where the victim of the harm has no agency in the deployment of the system producing harm. The \textbf{Sociotechnical Gap} is then the divide between a model's mathematical robustness and a system's resilience to actual harm. We propose that harm is the true objective function,
which is poorly estimated by current metrics. This misalignment results in our community misclassifying systems with no inherent real-world impact as high-risk, and those with high impact as low-risk.

Taking this view of risk and harm allows us to re-categorize other model observations, like hallucinations. Superficially, a hallucination represents a stochastic reliability failure, rather than a targeted adversarial intervention. However, from a systemic perspective, hallucinations constitute unforced vulnerabilities that an adversary can weaponize (e.g., indirect prompt injection attacks exploiting an LLM's propensity to hallucinate or follow untrusted context). Thus, while hallucinations originate from training distribution gaps, their operational risk is evaluated under the same systemic harm framework as intentional attacks.

We acknowledge that a broader community of researchers is already investigating the profound societal harms linked to AI misuse, systemic bias, and injustice \citep{nist2024adversarial, tan2022risks}. However, we believe the AML community should move in concert with these approaches, rather than tangentially, to better understand how technical failures can propagate through interconnected systems. This would form an exciting new research direction for understanding systemic resiliency.

\subsection{The Framing of Risk}

Crucially, the impacts of misalignment between research practice and real-world impacts relate not just to undiscovered vulnerabilities, but also to how risk is perceived, and how those perceptions affect behavior. By devoting the majority of research effort into the epsilon-ball, the AML community presents a narrative of risk that ignores particular interventions, particularly regarding data provenance. 

After all, the epsilon-ball is not just a metric; it is an articulation of safety for defenders, and a proxy of stealth for attackers, both of which can mislead stakeholders. When articulating risk in terms of norms---in terms of the size of an achievable attack or guaranteable defense---research encourages a perception of security that is divorced from production reality. A realistic articulation of safety would instead focus on the \textit{logic-to-action link}: at what point does a degraded model output trigger a catastrophic system state? If a production system requires a 95\% confidence threshold to take an action, an attack that shifts confidence from 99\% to 96\% may lead to success within a paper, but can be a non-event in reality. Of course, understanding such linkages once again implies a level of access to systems that researchers may not realistically be able to achieve. 

That is not to say that deployed systems are inscrutable. In informal interviews of Chief Information Security Officers and AI team leaders,
we have found that most treat data provenance and access restrictions as the key to practical AI security. These results are reinforced by studies demonstrating that corporate leadership are often unprepared in their knowledge of the risk landscape~\citep{kumar2020adversarial, boenisch2021never, ward2024ai}. Limiting our risk landscape to those that we can easily study fails those who would most benefit from our research.

\section{Alternative Views}

There exists an argument that $\ell_p$ is a deliberate simplification, and that we should not consider a broader risk landscape until we have resolved robustness in this space---that we should walk before we run. Our interest in this work is not to obviate all metrics, but to ensure that our metrics and developments are aligned with risk and harm---in that we should at least be walking in the right direction. If we consistently tell practitioners that $\ell_p$ is the only threat model that they should care about (or the only vector that we understand), then we engender a false sense of security, and \textbf{incentivize investments into forms of protection that are unlikely to provide real-world value}.

Another compelling view is that adversarial examples are \emph{not} a security threat, but instead reflect the existence of non-robust features in the data that can contain high levels of model accessible information content while remaining imperceptible to humans \citep{ilyas2019adversarial}. Under this paradigm, the epsilon-ball is not a proxy for risk, but a tool for understanding the fundamental difference between human and machine perception and learning. Even if adversarial attacks may be practically irrelevant for some deployment loops, they remain valuable for dissecting deep learning.

It may also be argued that adversarial training is necessary because it forces models to learn more human-aligned features, even if this comes at the cost of clean accuracy \citep{tsipras2019robustness}. From this perspective, the epsilon-ball is a regularizer that prevents a model from relying on spurious correlations. Improving $\ell_p$-robustness is thus about building more semantically consistent models---although this is not how this work is typically framed.

It could also be argued that the community is already beginning to pivot towards this more comprehensive view of risk, through work exploring the potential of forensically establishing provenance. While techniques like Deepfake Detection show promise, they still exhibit vulnerabilities to the very adversarial examples that we have argued against \citep{hussain2021adversarial}. With that said, the failure to be able to detect fakes underscores our points regarding the risks associated with generative examples. 

\section{A Pathway Forward}

Calling for the community to move beyond the orthodoxy of the $\epsilon$ ball is not an indictment of adversarial machine learning. Instead, we believe that it is an essential step for the maturation of our field. As researchers who have relied upon $\ell_p$ threat models ourselves, we recognize the mathematical elegance and theoretical clarity that these distance-bounded formulations have provided.  But we no longer operate in a space where AI applications are theoretical, and where harms are conceptual. Adversarially exposed AI systems are deployed all across the world, and their outcomes influence billions of people. In ignoring these realities, the AML community---which is motivated by practical goals---is eschewing any chance of impacting those that we seek to help. And the challenges here will only grow as synthetic data generation grows stronger. 

We believe that understanding and interrogating the limitations of current approaches is only half the battle. We therefore propose an actionable, three-pillar agenda to more strongly anchor adversarial robustness to both harm awareness and systemic safety. This proposal does not propose completely eschewing mathematical convenience, but rather suggests that we should  move towards closer alignment with real world practices, and, when dealing with abstractions, be more realistic about the impact of those abstractions on real world security. Being realistic, and truthful, about the impact of our work is crucial for the AML community delivering real-world impact. If we do not, we will end up with a community whose dominant presentation is that $\ell_p$ is the most important adversarial framework to consider, when it may well be the most meaningless.

\paragraph{Benchmarking} Our first pillar is that AML benchmarks should broaden beyond just considering different models and architectures, but to instead consider the breadth of the risk landscape, and the context that attacks and defences exist within. Audio, images, videos, for too many systems our observations are imperfect. Cameras introduce noise, audio observations must be processed through microphones, through transmission protocols, and through compression. These factors are often known, and sometimes acknowledged. But these systems do not just exist to take an input and pass it through to a model. These models exist within broader downstream contexts, which also affect the potential for risk and harm. 

We acknowledge that it would be impossible to construct benchmarks that fully articulate the totality of ways an input could be processed and passed to a model, let alone covering all possible downstream applications. However completely eschewing these factors should not be the answer either---finding a middle ground for evaluation and benchmarking should be a key direction for the community. 

Pursuing this would be valuable not just for measuring performance, but also for providing a framework for authors to critically analyze the real-world contexts that their works exist in. Elucidating and interrogating the assumptions inherent to our works is critical for understanding the problems that research is addressing. As the Empty Room Paradox shows, these factors are too often left unexplored, to the detriment of the community, and for our ability to impact the broader AI space. 

\paragraph{Cost Contextualization} As a part of our interrogation of AML practices, we believe that is crucial that attacks and defenses consider how we can quantify the socio-economic costs $C_{\text{attack}}$ associated with adversarial interventions. This would represents not the value to the attacker, but rather would serve as a comprehensive measure of what would be required for an attack to succeed: the query costs, infrastructure demands, and detection risks. This, to be clear, is an impossible task---researchers cannot reasonably test against broad swathes of hardware, nor keep track of purchasing costs which can change with both time and geography. However, using this as a conceptual framework for assessing the minimum systemic requirements for an attack to be successful would allow for the construction of a pareto frontier of risk. Rather than looking at the performance of a single (or small number of) attack(s), physical configurations, hardware setups, or detection patterns, risk should be considered in terms of the overall cost of an attack. 

\paragraph{Measuring operational impacts} The growing prevalance of AI in industrial contexts means that it should no longer be considered reasonable to abstract away the deployed reality of these systems. As AI becomes increasingly embedded into industrial workflows and safety-critical decisions, research must treat operational impacts and downstream harm as first-class evaluation criteria.

\subsection{Call for Action}

We believe the prima facie risk calculus of the AML community---which posits small-perturbation attacks as the primary threat---must be decisively reweighted. While constraints are necessary to define any research problem, the overemphasis on the small epsilon-ball has become a metrological ball-and-chain, dragging down our collective ability to communicate real-world risk. 

We call on the community to abandon the epsilon-ball as its default proxy for risk. True safety requires the community to align research with operational MLOps, adversarial economics, and downstream harms. Grounding research in these operational realities will both inspire interesting new research, and will ensure that adversarial machine learning is relevant to the security of the systems shaping our world.

\section*{Acknowledgments}

This work was supported by the Australian Defence Science and Technology (DST) Group via the Advanced Strategic Capabilities Accelerator (ASCA) program.

\bibliography{references}
\bibliographystyle{icml2026}

\appendix

\section{Metric Relativity}\label{app:relativity}

It is important to stress that small perturbations, as a concept, are fundamentally relativistic. The size of the perturbation will always be defined entirely by the metric chosen to measure it. A perturbation that is relatively large in one metric space can be perceptually negligible, where similarly another small metric may be perceptually catastrophic. This disconnect highlights the arbitrary nature of the constraints we place on adversarial attacks.

Consider a rotation of an image by one degree. In pixel space, this will alter the value of nearly every pixel in the frame, resulting in massive distances in almost any $\ell_p$ space~\citep{engstrom2019exploring}. Yet to a human observer, the image is perceptually equivalent. Similarly, global color transformations---such as inverting the color channels or converting an image to grayscale---represent enormous traversals across the input space. Despite this, the semantic content of the image remains perfectly preserved; a grayscale dog is still recognized instantly as a dog. These examples demonstrate that large pixel-space perturbations do not necessarily equate to semantic changes, challenging the assumption that risk is confined to a small radius around the input.

This metric relativity reinforces the \emph{Empty Room Paradox} introduced earlier. Just as an attacker in an empty room need not constrain their audio volume, an attacker operating in a domain where semantic preservation is robust to large shifts need not constrain themselves to a small $\ell_p$ ball. If a model is deployed in an environment where lighting conditions shift dramatically (which may appear as a large perturbation against many measures), an attacker can hide within those natural variations. By tethering our definition of risk to pixel-level proximity, we blind ourselves to the reality that adversarial influence can be exerted through large but semantically valid transformations that our current metrics classify as out-of-bounds or easily detectable, when in reality they are neither.

Advancing the field requires us to abandon the notion that safety is a geometric property measurable by a single norm. Instead, we must adopt a relativistic view where the allowable perturbation budget is defined not by mathematical distance, but by the specific semantic and operational invariances of the deployment environment. Only then can we move from defending against mathematical curiosities to defending against meaningful threats.

\section{Reinforcement Learning and the Access Fallacy}\label{app:rl}

Reinforcement Learning (RL) presents a particularly stark example of the how a disconnect between theoretical threat models and practical security can distort research practices. A significant portion of the adversarial RL literature operates under the assumption that an attacker can directly intercept and modify the state vector or observation passed from the environment to the agent~\citep{kiourti_trojdrl_2020, gong_mind_2022, yu_temporal-pattern_2022, chen2022marnet, wang2021stop, wang_backdoorl_2021, chen_backdoor_2022}. This access model allows for precise, mathematically calculated perturbations to be injected into the agent's perception or actions at every timestep. While this framework is intellectually rigorous and allows for the application of standard gradient-based attack methods, it rests on a foundational fragility: the assumption of access.

However, if an adversary possesses the ability to intercept and arbitrarily modify the raw memory tensors or data streams connecting a sensor to a decision-making unit, the system is already effectively compromised at the root level. An attacker with this level of privilege does not need to solve a complex optimization problem to slightly perturb a state vector to induce a suboptimal action; they could simply replace the action output, crash the system, or exfiltrate the reward function. Portraying the optimization of sophisticated exploits for systems that have already lost their most basic integrity must be considered to be a form of security theater.

It is important to acknowledge, however, that not all RL research falls into this trap. A subset of the literature correctly identifies that the true attack surface lies not in the internal memory bus, but in the external physical environment. Work that focuses on manipulating the physical observations---such as altering the visual field of a camera-based agent through real-world objects or light projection---avoids this access fallacy~\citep{gleave2019adversarial}. In these scenarios, the attacker legitimately controls only the external state, and the challenge of inducing failure through the system's own perception pipeline becomes a valid and dangerous threat model. Fundamentally, as is the case with all other data modalities, we must restrict ourselves to constraints that are not just mathematically convenient, but also reflect the physics of the overall event. 

\section{Autonomous Detection and Operational Constraints}\label{app:detection}

The transition from human-centric imperceptibility to machine-centric detectability is perhaps most visible in the context of autonomous network detection. In these environments, the Machine-in-the-Loop auditor does not view data through a lens of human visual perception, but rather by way of high-dimensional statistical flows and behavioral signatures. These works provide an example for the broader AML community, for they acknowledge the otherwise unacknowledged: that these systems will likely exist in a context where human perception is immaterial. 

In these systems, the relevant constraint for a real-world attacker is not minimizing a distance metric, but minimizing the probability of triggering a machine-side alarm across a range of potentially unknown statistical filters~\citep{metzen2017detecting}. This has allowed budgets to be more broadly aligned with socioeconomic factors, where every attempt to probe or manipulate a system carries an inherent cost in terms of compute resources, time-to-impact, and the risk of permanent exposure. If an attack requires ten thousand queries to converge on an adversarial example, the financial and operational cost of that search may well exceed the eventual payoff.

\section{Experimental Technique}\label{app:experimental}

Our paper analysis was based upon a multi-stag analytical process. We begun by employing the OpenReview API to collect accepted papers from ICML, ICLR, and NeurIPS over 2025--2025, the breakdown of which can be found in Table~\ref{tab:source_dist}, based upon the keywords `\{adversarial, robustness, poisoning, backdoor, trojan, jailbreak, prompt injection, unlearning, watermarking\}'. These papers were converted to rich text for searching. 

These papers were then categorized in a multi-stage process. We began by keyword searching the title and abstract to attempt to identify the data modality, the phase of the model life cycle that the attack was deployed against, and any constraints. Regex-based extraction then extracted out any sections labeled `\{Threat Model, Problem Formulation, Methodology, Evasions\}'. These were then parsed sequentially through Google Gemini. This sequential parsing was deliberate, as if all information was found within the Threat Model, then the additional sections were not provided to the LLM, in an attempt to minimize overall data ingestion. While it is likely that Gemini had previously been exposed to all of these works, we considered this minimal ingestion strategy to be the most appropriate to balance out concerns. For papers whose properties were not able to be parsed based upon this pass, the full paper was scanned for metrics, defenses, and constraints, which were used to build the classification.

Our process ensured accuracy through an additional two stage verification process. The first was that the LLM was not instructed to simply classify the model (based upon a number of categories), it was instructed to provide evidence, including direct quotes from the papers in question supporting these positions. These quotes were regex validated, to confirm that the quotes were not hallucinated. This process also allowed for papers to clearly be labeled as not providing sufficient quality information to appropriately classify, leading to $3208$ analyzed papers. We then pursued a blind human validation stage of $20$ randomly selected papers, where it was observed that both sets of results converged. 

\begin{table}
\caption{Distribution of Analyzed Papers by Venue and Year.}
\label{tab:source_dist}
\centering
\begin{small}
\begin{sc}
\begin{tabular}{ lrrr|r }
\toprule
Conference & 2023 & 2024 & 2025 & Total \\
\midrule
ICLR & 605 & 316 & 498 & 1419 \\
ICML & 251 & 341 & 470 & 1062 \\
NeurIPS & 453 & 569 & 808 & 1830 \\
\midrule
Total & 1309 & 1226 & 1776 & 4311 \\
\bottomrule
\end{tabular}
\end{sc}
\end{small}
\end{table}

\begin{table}
\caption{Data Poisoning by Mathematical Constraint Type.}
\label{tab:poison_math}
\centering
\begin{small}
\begin{sc}
\begin{tabular}{ lr }
\toprule
Mathematical Constraint & Count \\
\midrule
Unconstrained Content (Dirty Label) & 270 \\
Spatially Constrained (L0 / Mask) & 202 \\
Semantically Constrained (Clean Label) & 134 \\
Magnitude Constrained (Lp-Norm) & 112 \\
\bottomrule
\end{tabular}
\end{sc}
\end{small}
\end{table}

\begin{table}
\caption{Breakdown of Semantic/Transformation Constraints.}
\label{tab:semantic_math}
\centering
\begin{small}
\begin{sc}
\begin{tabular}{ lr }
\toprule
Mathematical Constraint & Count \\
\midrule
Distributional (Wasserstein/EMD) & 220 \\
General Semantic / Unspecified & 184 \\
Discrete Distance (Levenshtein) & 26 \\
Geometric (Rotation/Spatial) & 7 \\
\bottomrule
\end{tabular}
\end{sc}
\end{small}
\end{table}

\begin{table}
\caption{Breakdown of Physical/Patch Constraints.}
\label{tab:patch_math}
\centering
\begin{small}
\begin{sc}
\begin{tabular}{ lr }
\toprule
Mathematical Constraint & Count \\
\midrule
Spatially Constrained (L0 / Mask) & 91 \\
\bottomrule
\end{tabular}
\end{sc}
\end{small}
\end{table}

\begin{table}
\caption{Breakdown of LLM Jailbreak Constraints.}
\label{tab:jailbreak}
\centering
\begin{small}
\begin{sc}
\begin{tabular}{ lr }
\toprule
Optimization Constraint & Count \\
\midrule
Perplexity (Soft Constraint) & 605 \\
Perplexity + Semantic Embedding & 142 \\
Qualitative Stealth & 1 \\
\bottomrule
\end{tabular}
\end{sc}
\end{small}
\end{table}

\end{document}